\documentclass[a4paper,conference]{IEEEtran}
\IEEEoverridecommandlockouts

\usepackage[T1]{fontenc}

\usepackage{tikz}
\usepackage{cite}
\usepackage{amsmath,amssymb,amsfonts}
\usepackage{dsfont}
\usepackage{graphicx}
\usepackage{textcomp}
\usepackage{xcolor}
\usepackage{multirow}
\usepackage{booktabs}
\def\BibTeX{{\rm B\kern-.05em{\sc i\kern-.025em b}\kern-.08em
    T\kern-.1667em\lower.7ex\hbox{E}\kern-.125emX}}
\usepackage{fancybox}
\newcommand{\inda}{\phantom{1}\hspace{0.5mm}}
\newcommand{\indb}{\phantom{1}\hspace{3.5mm}}
\newcommand{\indc}{\phantom{1}\hspace{6.5mm}}
\newcommand{\indd}{\phantom{1}\hspace{9.5mm}}

\newcommand{\bW}{\boldsymbol{W}}
\newcommand{\bWo}{\boldsymbol{W}^\mathrm{o}}
\newcommand{\HAC}{{\tt{HAM}}}
\newcommand{\sHAC}{{\tt{sHAM}}}
\newcommand{\bx}{\boldsymbol{x}}

\newcommand{\bnz}{\boldsymbol{nz}}
\newcommand{\bcb}{\boldsymbol{cb}}
\newcommand{\bri}{\boldsymbol{ri}}

\makeatletter
\newcommand{\linebreakand}{%
  \end{@IEEEauthorhalign}
  \hfill\mbox{}\par
  \mbox{}\hfill\begin{@IEEEauthorhalign}
}
\makeatother

\begin{document}

\title{Compression strategies and space-conscious representations for deep neural networks}

\author{\IEEEauthorblockN{Giosuè Cataldo Marinò}
\IEEEauthorblockA{\textit{Università degli Studi di Milano} \\
Milano, Italy}
\and
\IEEEauthorblockN{Gregorio Ghidoli}
\IEEEauthorblockA{\textit{Università degli Studi di Milano} \\
Milano, Italy}
\linebreakand
\IEEEauthorblockN{Marco Frasca}
\IEEEauthorblockA{\textit{Università degli Studi di Milano} \\
Milano, Italy \\
marco.frasca@unimi.it}
\and
\IEEEauthorblockN{Dario Malchiodi}
\IEEEauthorblockA{\textit{Università degli Studi di Milano} \\
Milano, Italy \\
dario.malchiodi@unimi.it}
}

\maketitle

\begin{abstract}
Recent advances in deep learning have made available large, powerful convolutional neural networks (CNN) with state-of-the-art performance in several real-world applications. Unfortunately, these large-sized models have millions of parameters, thus they are not deployable on resource-limited platforms (e.g. where RAM is limited).
Compression of CNNs thereby
becomes a critical problem to achieve memory-efficient and possibly computationally faster model representations.
In this paper, we investigate the impact of lossy
compression of CNNs by weight pruning and quantization, and lossless weight matrix representations based on source coding. We  tested several combinations of these techniques on four benchmark datasets for classification and regression problems, achieving compression rates up to $165$ times, while preserving or improving the model performance.
\end{abstract}


\section{Introduction}
Although the main structural results behind deep neural networks (NN) date back to more than forty years ago, this field has constantly evolved, giving rise to original models, but also to novel techniques for controlling generalization error. The availability of powerful computing facilities and of massive amount of data allows nowadays to train extremely efficient neural predictors, setting the state-of-the-art in various fields, such as image processing or financial forecasting. Convolutional neural networks (CNN) play a prominent role: indeed, several pre-trained models, such as AlexNet~\cite{krizhevsky2012imagenet} and VGG16~\cite{Simonyan15}, are made available as a starting point for transfer learning techniques~\cite{transfer-learning}. These models, however, are characterized by a large memory footprint: for instance, VGG16 requires no less than 500 MB to be stored in main memory. As a consequence, querying such models becomes also demanding in terms of energy consumption. This clashes with the limitations of mobile phones, smartwatches, and in general with IoT-enabled devices.
Leaving aside the vein of \emph{compact model} production, aiming to directly induce succint CNNs~\cite{sandler2018mobilenetv2}, in this paper we focus on \emph{network compression}. Indeed, knowledge in a neural network is \emph{distributed} over millions, or even billions, of connection weights. This knowledge can be extracted transforming a learnt network into a smaller one, yet with comparable (or even better) performance. Main approaches proposed in the literature can be cast into the following four categories:
\begin{itemize}
    \item \emph{matrix decomposition}, aiming at writing over-informative weight matrices as the product of more compact matrices;
    \item \emph{data quantization}, focused on limiting the bitwidth of data encoding the mathematical objects behind a NN, such as weights, activations, errors, gradients, and so on;
    \item \emph{network sparsification}, aimed at reducing the number of free parameters of a NN, notably the connection weights;
    \item \emph{knowledge distillation}, consisting in subsuming a large network $L$ by a smaller one, trained with the target of mimicking the function learnt by $L$.
\end{itemize}
The reader interested to a thorough review of these methods can refer for instance to~\cite{deng2020model,cheng2017survey}. Note that we don't consider special techniques for convolutional layers, such as those tweaking the corresponding filters~\cite{zhai2016doubly}. This is due to the fact that, in most CNNs, the memory required by these layers is negligible w.r.t.\ that of fully-connected layers.

This work aims at investigating the joint effect of: (i) lossy compression for NN weights, and (ii) entropy coding, lossless  representation techniques
aware of the limited amount of memory  (aka \emph{space-conscious} techniques~\cite{Ferragina:2020pgm}). In particular, concerning matrix compression, we analyse the effectiveness of some existing pruning and quantization methods, and introduce a probabilistic quantization technique mutuated from federated learning. From the matrix storage standpoint, we propose a novel representation, named \textit{sparse Huffman Address Map compression} (\sHAC), combining entropy coding, address maps, and compressed sparse column (CSC) representations. \sHAC\ is specifically designed to exploit the sparseness and the quantization of the original weight matrix.
The proposed methods have been evaluated on two publicly available CNNs, and on four benchmarks for image classification and for the regression problem of drug-target affinity prediction, confirming that CNN compression can even improve the performance of uncompressed models, whereas the \sHAC\ representation can achieve compression rates up to around $200$ times. The work is organized as follows: Sects.~\ref{sec:compression-techniques} and \ref{sec:compressed-representations} describe the considered compression and representation techniques, while Sect.~\ref{sec:experiments} illustrates the above mentioned experimental comparison, depicted in terms of performance gain/degradation, achieved compression rate, and execution times. Some concluding remarks end the paper.


\section{Compression techniques}
\label{sec:compression-techniques}
This section describes three methodologies used in order to transform the matrix $\bWo \in \mathbb R^{n \times m}$ organizing the connection weights of one layer in a neural network into a new matrix $\bW$ which approximates $\bWo$, though having a structure exploitable by specific compression schemes (cfr.\ Sect.~\ref{sec:compressed-representations}) \footnote{these techniques can be applied separately to the dense layers of any NN.}. In the rest of the paper, $w^\mathrm{o}$ and $w$ will denote generic entries of $\bWo$ and $\bW$, respectively. Boldface and italic boldface will be used for matrices (e.g.\ $\bW$) and vectors ($\bx$), respectively, while $|\cdot|$ will be an abstract cardinality operator returning the length of a string or the number of elements of a vector. Finally, we define the \emph{sparsity coefficient} $s \in [0, 1]$ of $\bW$ as the ratio between the number of its nonzero elements and $mn$.

\subsection{Pruning}
Neural networks have several analogies with the human central nervous system. In particular, storing knowledge in a distributed fashion implies \emph{robustness} as a side effect: performance degrades gracefully when a damage occurs in the network components, i.e., when connections change their weight, or even get discarded. In turn, robustness can be exploited to compress a NN by removing connections which do not significantly affect the overall behaviour. This is referred to as \emph{pruning} a learnt neural network. An originally oversized NN might even be outperformed by its pruned version.

Pruning is typically done by considering connections whose weights are small in absolute value. Indeed, the signal processed by an activation function is computed as a weighted sum of the inputs to the corresponding neuron, precisely relying on connection weights. Thus, nullifying all negligible (positive or negative) weights should not sensibly change the above signal, as well as the network output. This is why we performed pruning by fixing an empirical percentile $w_p$ of the set of entries of $\bWo$, and subsequently defined the entries of $\bW$ setting $w = w^\mathrm{o}$ if $|w| > w_p$, $0$ otherwise. This procedure has a time complexity of $\mathcal O(nm \log (nm))$ (due to sorting).
As pruning has the effect of modifying the structure of the neural network, a post-processing phase retrains the network on the same data, now only updating non-null weights in $\bW$.
The only parameter is the percentile level $p$, which in turn is obviously related to the sparsity coefficient $s$ (see Sect.~\ref{sec:experiments} for a description of how $p$, as well as $k$ and $b$ in next sections, have been selected).

\subsection{Weight sharing}
When the weights in $\bWo$ assume a small number of distinct values,
applying the \emph{flyweight} pattern~\cite{gang-of-4}
results in a technique called \emph{weight sharing} (WS)~\cite{Han15}.
Distinct values are
stored in a table, whose indices are used as matrix entries. As (integer) indices
require less bits than (float) weights, the latter matrix is significantly compact and largely compensates for the
additional table\footnote{note that, in the original formulation, the representation of this matrix still scales with $mn$, while in Sect.~\ref{sec:experiments} we use a more efficient encoding.}. This comes at the price of requiring
two memory accesses in order to retrieve a weight.

Although $\bWo$
isn't expected to initially enjoy this property, robustness
is helpful in this case, too.
Close enough weights can be set to a common value
without significantly affecting network performance, yet allowing to apply WS. For instance, \cite{Han15}
clusterizes all $w^\mathrm{o}$ values,
setting $w$ to the
centroid of the corresponding $w^\mathrm{o}$. Assuming the $k$-means algorithm is used~\cite{mcqueen}, the time complexity is $\mathcal O(k(mn)^2)$,
where $k$ is
the number of different weight values.
A second retraining phase is advisable also here, though
updating weights is trickier, because the latter shall take values in the centroid set $\{c_1, \dots, c_k \}$. This is ensured using the cumulative gradient
\begin{equation*}
    \frac{\partial \mathcal L}{\partial c_l} = \sum_{i,j} \frac{\partial \mathcal L}{\partial w_{ij}}\mathds{1} (I_{ij}=l),
\end{equation*}
where $l \in \{ 1, \dots, k \}$, $I_{ij}$ is the cluster index of $w^\mathrm{o}_{ij}$, and $\mathds{1}$ is the indicator function. Applying cumulative gradient might end up in using less than $k$ distinct weights, if some representatives converge to a same value during retraining.
Pruning and weight sharing can be applied in chain, with weight sharing only considering the non-null weights identified by pruning.

\subsection{Probabilistic quantization}
A recent trend on quantization
relies on probabilistic projections of weight onto special binary~\cite{NIPS2015_5647} or ternary values~\cite{deng2018gxnor}. Here we present an alternative approach, named \textit{Probabilistic Quantization} (PQ), mutuated and extended from the federated learning context~\cite{federated-learning}, and never used for NN compression. PQ is based on
the following probabilistic rationale.
Let $\underline w$ and $\overline w$ denote the minimum and maximum weight in $\bWo$, respectively, and suppose that each learnt weight $w^\mathrm{o}$ is the specification of a random variable $W^\mathrm{o}$ distributed according to a fixed, yet unknown, distribution having $\mathcal W := [\underline w, \overline w]$ as support. Let $W$ be the two-valued random variable defined by $\mathrm P(W = \underline w) = \frac{\overline w - w}{\overline w - \underline w}$ and $\mathrm P(W = \overline w) = \frac{w - \underline w}{\overline w - \underline w}$. The observations of $W$ approximate a weight $w$ through an extreme form of WS (cfr.\ previous section), using an approach different from k-means for finding representative weights. Now, $\mathcal E(W | W^\mathrm{o}=w) = w$, and, in turn, $\mathcal E(W) = \mathcal E(W^\mathrm{o})$ regardless how $W^\mathrm{o}$ is distributed. Thus, \emph{simulating} $W^\mathrm{o}$ for each entry $w^\mathrm{o}$
we obtain an approximation $\bW$ of $\bWo$ having the desirable \emph{unbiasedness property} that the two corresponding random matrices have the same expected value.
This method has been heuristically extended by partitioning $\mathcal W$ in $b > 2$ intervals. A generic $w^\mathrm{o}$ is compressed precisely as in the two-values case, but now $\overline w$ and $\underline w$ denote now the extremes of the interval containing $w^\mathrm{o}$.
We remark here that sub-intervals, and therefore representative weights, can be chosen in order to preserve the above mentioned unbiasedness property.
Indeed,
this happens when the intervals' extremes are
$\chi_{\frac{i}{b}}$, for $i = 1, \dots, b$, where $\chi_q$ denotes the $q$-quantile of $W$\footnote{this requires the weights to follow a common probability distribution; however, no additional hypotheses are needed.}. The time complexity of the overall operation is $\mathcal O(nm \log (nm))$ (due to quantile computation).
Note that the same considerations pointed out for tWS at the end of previous section, namely retraining via cumulative gradient formula and combined use with pruning, also apply to PQ.

\section{Compressed Matrix Representation}
\label{sec:compressed-representations}
The matrix $\bW$ obtained using any of the techniques described in Sect.~\ref{sec:compression-techniques} has as many elements as the original matrix. However, it exhibits properties
exploitable by a clever encoding, so that $\bW$ is stored using less than $\mathcal O(mn)$ memory locations, as required by the classical row-order method.
In this section, two existing compressed representations of $\bW$ are first described, then a novel
method is proposed, overcoming their limitations and explicitly profiting from sparsity and presence of repeated values. Moreover, the method
does not require assumptions on the matrix sparsity, on the distribution of nonzero elements, or on the presence of repeated values.
To be ablle to efficiently compute the dot product $\bx^T\bW$, where $\bx \in \mathbb{R}^{n\times 1}$, necessary for the forward computation in a NN, a dedicated
procedure is also described.

\subsection{Compressed sparse column}\label{sub:CSC}
  The \textit{compressed sparse column} (CSC) format \cite{CSC} is a common general storage format for sparse matrices. It is composed of three arrays:
\begin{itemize}
 \item[-] $\bnz$, containing the nonzero values, listed by columns;
 \item[-] $\bri$, containing the row indices of elements in $\bnz$;
 \item[-] $\bcb$, where the difference $cb_{i+1}-cb_i$ provides the number of nonzero elements in column $i$; thus, $\bcb$ has dimension $m+1$, where $cb_{m+1}=cb_1 + |\bnz|$.
\end{itemize}
As an example, consider the matrix

\begin{equation}\label{eq:mat_ex}
\bW =
\begin{pmatrix}
1 & 0 & 4 & 0 & 0\\
0 & 10 & 0 & 0 & 0\\
2 & 3 & 0 & 0 & 5\\
0 & 0 & 0 & 0 & 0\\
0 & 0 & 0 & 0 & 6\\
\end{pmatrix},
\end{equation}
whose corresponding CSC representation is $\bnz = \begin{pmatrix}1& 2 & 10& 3& 4& 5& 6\end{pmatrix}$, $\bri =  \begin{pmatrix}1& 3& 2& 3& 1& 3& 5\end{pmatrix}$, and $\bcb =  \begin{pmatrix}1& 3& 5& 6& 6& 8\end{pmatrix}$.
Let $q = |\bnz|$ be the number of nonzero elements in $\bW$, and $B$ be the number of bits used to represent every element of the matrix (one memory word), so that we need $Bnm$ bits to store $\bW$, and $(2q+m+1)B$ to store its CSC representation. Note that we assumed $B$ bits are needed also for the components of $\bri$, although they can be represented using only ${\lceil\log n\rceil}$ bits, which might be lower than $B$. Thus the occupancy proportion is given by $\psi_{CSC} = \frac{2q+m+1}{nm}$.

Denoting by $s$ the sparsity coefficient of $\bW$, we have $q=snm$, thus $\psi_{CSC}<1$ implies $s < \frac{1}{2}$.
The matrix dot product $\bx^T\bW$ will be computed through the typical dot product for CSC format, with computational complexity $\mathcal{O}(q)$ \cite{CSC}, that can be sped up through parallel computing.
The main limitation of CSC is tye use of $B$ bits for any elements of the matrix, whereas variable length coding can provide more compact representations and higher bit-memory efficiency.
 \subsection{Huffman address map compression}
  The idea of using Huffman coding after network pruning and quantization is introduced in \cite{Han15}, although with little detail. Here, in addition to provide an exhaustive description, we also point out a main limitation of this approach: it does not directly profit from the matrix sparsity.
 Like CSC, this is a lossless compression technique, based on Huffman coding and the address map logic \cite{Pooch73}, which we named \textit{Huffman Address Map compression} (\HAC).
In address maps, matrix elements are treated as a sequence of bits, concatenated by rows or by columns, where null entries correspond to the bit $0$, and each nonzero element $z$ is substituted by a binary string encoding its address $a(z)$, and concatenated to the rest of the stream. For instance, the bit stream for the matrix defined in (\ref{eq:mat_ex}) is
\begin{displaymath}
    a(1) 0 a(2) 0 0 0 a(10)a(3)00a(4)00000000000a(5)0a(6)\ .
\end{displaymath}
 To be efficient, this storage needs a compact representation of addresses.  
 The Huffman coding $H_{\bW}(z)$ of nonzero values $z$ is a uniquely decodable and instantaneous code ensuring a near-optimal compression rate \cite{Huffman52}. Indeed, given a source $(w_1, \ldots, w_l)$ whose symbols have probabilities $(p_1, \ldots, p_l)$, the average number of bits per symbol $\overline H_{\bW} := \sum_{i=1}^l p_i |H_{\bW}(w_i)|$ is almost equal to the optimal value $\mathcal{H}=-\sum_{i=1}^l p_i \log p_i$ corresponding to the entropy of the source (when the symbols are independent and identically distributed). More precisely,  $\mathcal{H} \leq |\overline H_{\bW}| \leq \mathcal{H} + 1$, and $\mathcal{H}$  corresponds to the minimal average number of bits per symbol, according to Shannon's source coding theorem \cite{Shannon48}.


Once the Huffman code $H_{\bW}$ has been built, we replace each $a(z)$ in the bit stream with the corresponding $H_{\bW}(z)$.
In order to have a uniquely decodable string, 
zeros are also included in the Huffman code, thus having a total of $q+1$ codewords. The resulting bit stream
$\HAC(\bW)$ is then split into $N=\lceil \frac{|\HAC(\bW)|}{B}\rceil$ memory words, $\HAC(\bW)_1, \ldots, \HAC(\bW)_N$, represented as an array $\mathcal{C}_{\HAC}(\bW)$ of $N$ unsigned integers. If $|\HAC(\bW)|$ is not a multiple of $B$, zero-padding is added to the last word.

To estimate $|\HAC(\bW)|$, we can assume the worst case for the entropy value, that is when all symbols are distinct ($nm$ symbols appearing exactly once in the matrix): in this case
$\mathcal{H} = \log (nm)$, and the Huffman code has an average codeword length upper-bounded by $1+\log (nm)$, which implies at most $nm(1+\log(nm))$  bits are needed.
To store $H_{\bW}$, and its inverse $H_{\bW}^{-1}$ used to decode, we need extra space: although there are methods storing a $n$-symbols Huffman code using at most $\lceil 10.75n\rceil -3$ bits \cite{Sultana12}, to ensure optimal search time, a `classical'  B-tree representation is used for both $H_{\bW}$ and $H_{\bW}^{-1}$, being aware that space occupancy can be improved. Assuming each value is represented trough 1
word ($B$ bits), each dictionary requires $3(q+1)B$ bits, $2B$ to store each pair $z$ and $H(z)$, and $B$ bits to store a pointer in the B-tree structure---overestimated, since we have less pointers than keys in a B-tree.
Overall, \HAC\ requirements are upper-bounded by $nm(1+\log(nm)) + 6mnB$ bits, which is more than $mnB$ bits required by an uncompressed matrix. For this reason, in the experiments only using pruning, the CSC representation is adopted.
As opposite, the space occupancy of \HAC\
decreases when only $k$ distinct weights are present in $\bW$, like in the output of WS and PQ (see Sections \ref{sec:compression-techniques}). Indeed, in the worst case (all symbols are equally probable, entropy $\log k$), the occupancy proportion is at most $\psi_{\HAC} = \frac{1+\log k}{B} + \frac{6k}{nm}$,
where as expected, for small $k$ the first term
is more relevant, while the second term grows faster with $k$.

\noindent\textit{Dot product}.
The procedure {\tt Dot$_{\HAC}$} (Fig.~\ref{fig:pseudocode.HAC}) executes the dot product $\bx^T\bW$, when $\bW$ is represented through the \HAC\ format.
It processes one compressed word of $\mathcal{C}_{\HAC}(\bW)$ at a time, obtains its binary representation $S$ (line $3$),
which is scanned at lines $4$-$10$ to detect code words. The procedure NCW gets the next code word from $S$, starting at the current bitstring offset $oset$,
possibly adding at the beginning of $S$ the bits $rem$ remaining from previous word.
If, starting from the current offset, no code word can be detected in $S$ (NCW returns $null$), it means that the next code word has been split on two adjacent memory words, accordingly $rem$ is updated and the next word will be read in the next iteration at line $2$. The procedure also takes into account for the $0$ padding.
Then,  the  weight relative to the code word detected is computed (line $5$), and multiplied by the corresponding element of $\bx$, to update the cumulative sum stored in variable $sum$, thus requiring to keep in memory only one weight at a time.

In summary, the $N$ iterations take time $\mathcal{O}(NB)=\mathcal{O}(nm)$  for line $3$, $\mathcal{O}(N)$ for lines $5$-$8$, and $\mathcal{O}(nm\log k)$ for line $4$, leading to an overall time complexity $\mathcal{O}(nm\log k)$.

\begin{figure}[ht]
\cornersize*{0pt}
{\hspace{-0.22cm}
\begin{minipage}{0.5\textwidth}
\caption{Pseudocode of the dot procedure for \HAC\ representation.}
\label{fig:pseudocode.HAC}
Procedure {\tt Dot$_{\HAC}$}\\
\noindent
{\tt Input}:
compressed array $\mathcal{C}_{\HAC}(\bW)$;
decoding dictionary $H_{\bW}^{-1}$;
vector $\bx \in \mathbb{R}^{n\times 1}$;
number of compressed words $N$;
\bigskip

\noindent
{\tt begin algorithm}\\
01:\inda   Initialize: $out:=$zeros$(n)$, $row:=1$, $col:=1$\\
\indc\indc\indb $sum:=0$, $rem:=null$, $oset:=0$\\
02:\inda   {\bf for each} $i$ {\bf from}  $1$ {\bf to} $N$ {\bf do} \\
03:	\indb   $S := $ getBinarySeq($\mathcal{C}_{\HAC}(\bW)[i]$)\\
04:	\indb  {\bf while} $[oset, rem, z]:=$ NCW($S, rem, oset)\neq null$\\
\indd{\bf do}\\
05:    \indc  $sum:=sum+x[row]*H_{\bW}^{-1}(z),\ row:=row+1$\\
06:    \indc  {\bf if} $row>n$ {\bf then} \\
07:    		\indd  $row:=1,\ out[col] := sum$\\
08:			\indd  $col:=col+1,\ sum := 0$\\
09:	 	\indc {\bf end if}\\
10:	\indb {\bf end while}\\
11:\inda {\bf end for}\\
{\tt end algorithm}\\
\noindent
{\tt Output}: $out$, that is $\bx^T \bW$.
\end{minipage}
}
\end{figure}
 \subsection{Sparse Huffman address map compression}\label{sec:sham}
One drawback of \HAC\ representation is that it does not directly exploit the sparsity of the matrix, which only indirectly induces a reduction in the space occupancy, due to the more compact resulting Huffman code (symbol $0$  has high frequency). When the matrix is large and very sparse, even using only $1$ bit to represent the symbol $0$, much memory would be required (e.g. $10s$ GB for a $10^5\times 10^5$ matrix).
To address this issue, the novel \textit{sparse Huffman Address Map compression} (\sHAC) is proposed, extending the \HAC\ format as follows.
The symbol $0$ is excluded from the bit stream and from the Huffman code, and a bitwise CSC representation of the matrix is adopted, producing the vectors $\bnz, \bri, \bcb$ (cfr.\ Sect.~\ref{sub:CSC}), but storing $\bnz$ using the $\HAC$ format. Namely, the Huffman code $H_{\bnz}$ for nonzero elements is built, and the corresponding bit stream $\sHAC(\bnz)= H_{\bnz}(nz[1])\ldots H_{\bnz}(nz[q])$ is obtained by concatenating their Huffman coding. $\sHAC(\bnz)$ is stored in the array  $\mathcal{C}_{\sHAC}(\bnz)$ of $N_1=\lceil \frac{|\sHAC(\bnz)|}{B}\rceil$ memory words. Considering the worst case, in which all $q=snm$ symbols are distinct, $\mathcal{C}_{\sHAC}(\bnz)$ is composed of $snm(1+\log snm)$ bits, in addition to the $6snmB$ for the dictionaries, and to the $B(n+m+1)$ bits required for $\bri$ and $\bcb$. The resulting occupancy ratio is $\psi_{\sHAC} = \frac{s(1+\log snm)}{B} + 6s + \frac{n+m+1}{nm}$.

On the other side, when only $k$ distinct values are present in $\bnz$, and again assuming the worst case for the Huffman coding, the occupancy ratio becomes $\psi_{\sHAC} = \frac{s(1+\log k)}{B} + \frac{6k}{nm} + \frac{n+m+1}{nm}$, where first term on the right is scaled by $s$ w.r.t. $\psi_{\HAC}$,  emphasizing the gain of increasing the sparsity of $\bW$,
whereas last term is constant w.r.t.\ $k$ and $s$.
Thus, when $s$ is such that $\frac{s(1+\log k)}{B} + \frac{n+m+1}{nm} < \frac{(1+\log k)}{B}$, it follows $\psi_{\sHAC} < \psi_{\HAC}$.

\noindent{\textit{Dot product}}.
Figure \ref{fig:pseudocode.sHAC} describes
the procedure executing the dot product $\bx^T\bW$ when $\bW$ is represented through the \sHAC\ format.
It extracts in sequence the compressed words of $\mathcal{C}_{\sHAC}(\bnz)$, computes the corresponding binary representation $S$ (line $3$), and  executes lines $4$-$13$ to detect code words. NCW is the same procedure used for {\tt Dot$_{\sHAC}$}, whereas the cycle at lines $5$-$7$ 
possibly skips empty columns.
The variable $pos$ contains the position in $\bnz$ of the current element $z$.
Line $8$ finds the weight relative to the code word detected, multiplies it by the corresponding element of $\bx$, and updates the column cumulative sum stored in variable $sum$.
The $N_1$ iterations require $\mathcal{O}(N_1B)= \mathcal{O}(snm)$
steps for line $3$, $\mathcal{O}(N_1)$ for lines $8$-$12$, $\mathcal{O}(m)$ for cycle $5$-$7$, and $\mathcal{O}(snm\log k)$ for line $4$. The overall time complexity is thereby $\mathcal{O}(snm\log k)$.

\begin{figure}[ht]
\cornersize*{0pt}
{
\begin{minipage}{0.5\textwidth}
\caption{Pseudocode of the dot procedure for \sHAC\ representation.}
\label{fig:pseudocode.sHAC}
Procedure {\tt Dot$_{\sHAC}$}\\
\noindent
{\tt Input}:
compressed array $\mathcal{C}_{\sHAC}(\bnz)$;
row index vector $\bri$;
vector $\bcb$;
vector $\bx \in \mathbb{R}^{n\times 1}$;
decoding dictionary $H_{\bnz}^{-1}$;
number of compressed words $N_1$;
\bigskip

\noindent
{\tt begin algorithm}\\
01:\inda   Initialize: $out:=$zeros$(n)$, $pos:=1$, $col:=1$\\
\indc\indc\indb $sum:=0$, $rem:=null$, $oset:=0$\\
02:\inda   {\bf for each} $i$ {\bf from}  $1$ {\bf to} $N_1$ {\bf do} \\
03:	\indb   $S := $ getBinarySeq($\mathcal{C}_{\sHAC}(\bnz)[i]$)\\
04:	\indb  {\bf while} $[rem, oset, z]:=$ NCW($S, rem, oset)\neq null$ {\bf do}\\
05:	   \indc  {\bf while} $cb[col+1] = pos$ {\bf do} \\
06:    	 \indd  $col := col + 1$, $out[col] := 0$\\
07:	   \indc {\bf end while}\\
08:    \indc  $sum:=sum+x[ri[pos]]*H_{\bnz}^{-1}(z)$,
        $pos:= pos+1$\\
09:    \indc  {\bf if} $cb[col+1] = pos$ {\bf then} \\
10:    		\indd  $out[col] := sum$\\
11:			\indd  $sum := 0,\ col := col +1$\\
12:	 	\indc {\bf end if}\\
13:	\indb {\bf end while}\\
14:\inda {\bf end for}\\
{\tt end algorithm}\\
\noindent
{\tt Output}: $out$, that is $\bx^T \bW$.
\end{minipage}
}
\end{figure}

The procedure {\tt Dot$_{\sHAC}$}   can be adapted to parallel computation by substituting
$\bcb$ with the vector containing the beginning of each column in the bitstream $\sHAC(\bnz)$, and using the current position in the bitstream ($oset$) to detect the end of columns. In this way, each column product can be run 
in parallel (e.g., through GPU);
we considered this as a future development, since in the empirical evaluation the
sequential version
was
very close to the full matrix parallel dot product testing time, on sufficiently sparse and quantized matrices. Analogously,
also {\tt Dot$_{\HAC}$}
can be parallelized.
\section{Experiments and Results}\label{sec:experiments}
To assess the quality of the proposed techniques, an empirical evaluation has been carried out on four datasets and two uncompressed models, as explained here below.

\subsection{Data}
\begin{itemize}
    \item \textit{Classification}.
    The MNIST database \cite{MNIST} is a classical large database of handwritten digits, containing 60K+10K  28x28 grayscale images (train and test set, respectively) from 10 classes (digits 0-9). 
    The CIFAR-10 dataset~\cite{Krizhevsky09learningmultiple} consists of 50K+10K 32x32 color images belonging to 10 different classes. 
    Both datasets are balanced w.r.t.\ labels.
    \item \textit{Regression}. We predicted the affinity between drug (ligand) and targets (proteins)~\cite{DeepDTA}, using the DAVIS~\cite{Davis11} and KIBA~\cite{KIBA14} datasets.
    Proteins and ligands are both represented through strings, respectively using the amino acid sequence and the SMILES (Simplified Molecular Input Line Entry System) representation. DAVIS and KIBA contain, respectively, $442$ and $229$ proteins, $68$ and $2111$ ligands, $30056$ and  $118254$ total interactions.
\end{itemize}
\subsection{Benchmark models}
 To have a fair comparison of the various compression techniques, we selected top-performing
 CNN models publicly available: (i) \textit{VGG19} \cite{Simonyan15},
 made up by $16$ convolutional layers and a fully-connected block (two hidden layers of $4096$ neurons each, and a softmax output layer)\footnote{Source code:  {\tt https://github.com/BIGBALLON/cifar-10-cnn}}, trained on CIFAR-10 and MNIST datasets; and (ii) \textit{DeepDTA} \cite{DeepDTA},  with distinct convolutional blocks for proteins and ligands (each composed of $3$ convolutional and a MaxPool layers), combined in a fully connected block consisting of 3 hidden layers of $1024$, $1024$, $512$ units, and a single-neuron output layer\footnote{Source code: {\tt https://github.com/hkmztrk/DeepDTA}}.

The original work using DeepDTA operated a $5$-fold cross validation (CV) to perform model selection, thus training on 4/5 of available data.
We retained the best configuration for hyperparameters and trained the CNN on the entire training set, leaving unchanged the original settings.
\subsection{Evaluation metrics}\label{sub:metrics}
We considered the difference $\Delta_{perf}$ between performances of compressed and uncompressed models, the ratio of testing time of the uncompressed model w.r.t.\ the compressed one (named $time$), and the space occupancy ratio $\psi$ (cfr.\ Sect.~ \ref{sub:CSC}).
As in the original works, we computed performance using \textit{Accuracy} and \textit{mean squared error} (MSE) for classification and regression, respectively.
Time and space performance account only for the actually compressed weights, that is those in fully-connected layers.
Moreover, the implementation of dot product for uncompressed models
exploits parallel computations implemented in Python,
thus penalizing $\HAC$ and $\sHAC$,
implemented sequentially
(their parallelization is planned as an extension). As shown in the next section, even with this penalty, our method approaches the uncompressed time when the matrix is sufficiently sparse and quantized.
\subsection{Compression techniques setup}
We tested all
combinations of
Pruning (Pr), WS, PQ, Pr-WS, Pr-PQ, selecting hyper-parameters as follows.
\begin{itemize}
\item \textit{Pruning}. We tested percentiles with $p \in \{30,  40,  50, \linebreak 60,  70,  80,  90,  95,  96,  97,  98, 99\}$; values $30$ and $40$ (for which CSC  does not achieve any compression) are included, as potentially useful in Pr-WS and Pr-PQ.
\item \textit{WS}. For VGG19, $k = 2, 32, 128, 1024$ was tested in the first two hidden layers $k = 2, 32, 128, 1024$, as well as $k = 2, 32$ in the third one (which is smaller).
DeepDTA is smaller than VGG19, so
all combinations of $k=2, 32, 128$ have been tested in the hidden layers, and of $k=2, 32$ for the output layer, due to its dimension.
\item \textit{PQ}. In order to have a fair comparison, $b$ took the same values as $k$ in the WS procedure;
\item \textit{Pr-X}. The combined application of pruning followed by the quantization $X \in \{WS, PQ\}$, was tested in two variants: a)
best $p$ in terms of $\Delta_{perf}$
is selected, and the parameters for $X$ are subsequently tuned a in previous points, and
b) vice-versa.
\end{itemize}

\noindent {\textit {Fine tuning of compressed weights}.}
The same configuration of original training procedure has been kept for the retraining after compression.
Data-based tuning was applied only to learning rate after retraining ($3\cdot10^{-4}$ for pruning, $10^{-3}$ and $10^{-4}$ for PQ, WS, and combined schemes), and the maximum number of epochs, set to  $100$.
\subsection{Software implementation}
The source code retrieved for baseline NNs was implemented in Python, using the Tensorflow
and Keras
libraries. Compression techniques and retraining procedures have been implemented in Python as well, also exploting GPUs. 

\subsection{Results}
As baseline comparison, Table~\ref{original_res} reports the testing results of the uncompressed models. The top performing results for each compression technique, along with the corresponding configuration, are shown in Table \ref{best_perf_compr}. To also evaluate compression capability, Table \ref{best_space_compr} contains the least occupying configuration for each compression methods having performance greater or equal to the original model (when available). Weight quantization is more accurate than pruning for classification, with PQ and WS on having the top absolute performance on MNIST and CIFAR-10, respectively. Overall,
 all techniques outperform the baseline, while exhibiting remarkable compression rates.
Similar trends raise for regression, where however pruning top-performs, and where PQ never improves the baseline MSE on KIBA dataset.
Performance improvements are particularly remarkable on DAVIS data (till around $30\%$ of baseline).
As expectable, the largest compression (while preserving accuracy) is achieved on the bigger net, VGG19, with a compression rate of more than $150$ times on CIFAR-10, with Pr-PQ and \sHAC\ representation.
Anyway, Pr-PQ method improves the baseline MSE of $ 17.1\% $, while compressing around $18$ times, also for DeepDTA.
\begin{table}[t]
\caption{Testing performance of original uncompressed models. Performance shows  accuracy for MNIST and CIFAR-10, and  MSE for  KIBA and DAVIS. \textit{Time} is the overall testing time.}
\begin{center}
\begin{tabular}{|l|l|c|c|}
\hline
                \textbf{Net} & \textbf{Dataset} & \textbf{Performance} & \textbf{Time (s)}\\ \hline
\multirow{2}{*}{VGG19} & MNIST & $0.9954$ & $8.88\cdot10^{-1}$ \\ \cline{2-4}
                  & CIFAR10 & $0.9344$ & $8.97\cdot10^{-1}$ \\ \hline
\multirow{2}{*}{DeepDTA} & KIBA & $0.1756$ & $1.75\cdot10^{-1}$ \\ \cline{2-4}
                  & DAVIS & $0.3223$ & $4.00\cdot10^{-2}$ \\ \hline
\end{tabular}
\end{center}
\label{original_res}
\end{table}
\begin{table}
\centering
\caption{Top testing performance achieved by compression techniques. \textit{Type} is the compression technique, while \textit{Perf} contains Accuracy for VGG19 and MSE for DeepDTA. \textit{$\psi$} is the occupancy ratio, whereas $*$ denotes \sHAC\ representation as the lowest occupancy on that setting (w.r.t. \HAC). In bold the best results on each couple Net-Dataset.}
\label{best_perf_compr}
\begin{tabular}{|l|c|c|c|c|}
\toprule
Net-Dataset & Type & Configuration & Perf & $\psi$ \\
\hline
\multirow{7}{*}{VGG19-MNIST}   & Pr & 96 & 0.9954   & 0.08     \\
\cline{2-5}
  & WS &  128-32-32  & 0.9957  & 0.321  \\
\cline{2-5}
  & PQ  &  32-32-2  & \textbf{0.9958}  & 0.309  \\
\cline{2-5}
  & Pr-WS a &  96/128-32-32   & 0.9956  & 0.039*  \\
\cline{2-5}
  & Pr-WS a &  96/128-32-32   & 0.9956  & 0.039*  \\
\cline{2-5}
  & Pr-PQ a &  96/32-128-32 & 0.9956  & \textbf{0.026}*  \\
\cline{2-5}
  & Pr-PQ b &  50/32-32-2     & \textbf{0.9958}  & 0.187   \\
\hline
\multirow{7}{*}{VGG19-CIFAR10} & Pr    & 60    & 0.9365   & 0.8     \\
\cline{2-5}
   & WS      &  32-32-2   & \textbf{0.9371} & 0.306  \\
\cline{2-5}
   & PQ      &  32-2-32   & 0.9363 & 0.091  \\
\cline{2-5}
  & Pr-WS a  &  60/2-2-32 & 0.9366 & 0.088  \\
\cline{2-5}
  & Pr-WS b  & 50/32-32-2 & 0.9370 & 0.216  \\
\cline{2-5}
  & Pr-PQ a  & 60/2-2-32  & 0.9363 & 0.088  \\
\cline{2-5}
  & Pr-PQ b &  98/32-2-32 & 0.9365 & \textbf{0.012}*  \\
\hline
\multirow{7}{*}{DeepDTA-KIBA}  & Pr    & 60    & \textbf{0.1599}  & 0.8  \\
\cline{2-5}
  & WS   & 128-128-32-2  & 0.1679  &    0.390    \\
\cline{2-5}
  & PQ   & 32-128-128-32   & 0.1761  &   0.425     \\
\cline{2-5}
  & Pr-WS a  & 60/32-128-2-32   &   0.1666  & \textbf{0.187}       \\
\cline{2-5}
  & Pr-WS b  &  30/128-128-32-2  &  0.1644   & 0.33       \\
\cline{2-5}
  & Pr-PQ a  & 60/128-128-128-32   & 0.1769     &    0.207    \\
\cline{2-5}
  & Pr-PQ b &  40/32-128-128-32   & 0.1683    &    0.291    \\
\hline
\multirow{7}{*}{DeepDTA-DAVIS} & P    & 80    & \textbf{0.2242}  & 0.4     \\
\cline{2-5}
  & WS   & 128-2-128-2   & 0.2320 & 0.212       \\
\cline{2-5}
  & PQ &  128-32-32-32    & 0.2430   &  0.324           \\
\cline{2-5}
  & Pr-WS a  &  80/32-128-2-32  & 0.2341 &  \textbf{0.105}  \\
\cline{2-5}
  & Pr-WS b  & 40/128-2-128-2   &  0.2826   & 0.191       \\
\cline{2-5}
  & Pr-PQ a  & 80/128-128-32   &    0.2302 &  0.122      \\
\cline{2-5}
  & Pr-PQ b &  60/128-32-32-32   &  0.2353   &   0.160     \\
\bottomrule
\end{tabular}
\end{table}
\begin{table}
\centering
\caption{Best occupancy ratio ensuring no decay in performance w.r.t.\ uncompressed model. Same notations as in Table~\ref{best_perf_compr}.
}
\label{best_space_compr}
\begin{tabular}{|l|c|c|c|c|}
\toprule
Net-Dataset                    & Type & Configuration & Perf     & $\psi$  \\
\hline
\multirow{7}{*}{VGG19-MNIST}   & Pr    & 97            & 0.9953   & 0.06     \\
\cline{2-5}
 & WS   & 128-2-32     & 0.9954   & 0.104  \\
\cline{2-5}
& PQ   & 1024-2-32       & 0.9955   & 0.126  \\
\cline{2-5}
& Pr-WS a  & 96/2-128-2 & 0.9954   & 0.038*  \\
\cline{2-5}
& Pr-WS b & 96/128-32-32 & \textbf{0.9956}   & 0.039*  \\
\cline{2-5}
& Pr-PQ a  & 96/32-128-32 & \textbf{0.9956}   & 0.026*  \\
\cline{2-5}
& Pr-PQ b & 97/32-32-2    & 0.9955   & \textbf{0.018}*   \\
\hline
\multirow{7}{*}{VGG19-CIFAR10} & Pr    & 99            & 0.9357   & 0.02     \\
\cline{2-5}
& WS   & 2-2-32       & 0.9360   & 0.063  \\
\cline{2-5}
& PQ   & 2-2-32        & 0.9351   & 0.063  \\
\cline{2-5}
& Pr-WS a  & 60/2-2-32    & \textbf{0.9366}   & 0.088  \\
\cline{2-5}
& Pr-WS b & 99/32-32-2   & 0.9358   &  \textbf{0.006}*  \\
\cline{2-5}
& Pr-PQ a  & 60/2-2-32    & 0.9363   & 0.088  \\
\cline{2-5}
& Pr-PQ b & 99/32-2-32   & 0.9363   & \textbf{0.006}*  \\
\hline
\multirow{7}{*}{DeepDTA-KIBA}  & Pr    & 60            & \textbf{0.1599}  & 0.8     \\
\cline{2-5}
& WS   & 32-32-2-2  & 0.1723  & 0.228       \\
\cline{2-5}
& PQ   & 32-128-128-32   & 0.1761  &   0.425      \\
\cline{2-5}
& Pr-WS a  &  60/32-2-32-2    & 0.1739          &   \textbf{0.127}     \\
\cline{2-5}
& Pr-WS b & 60/128-128-32-2  & 0.1712     & 0.222       \\
\cline{2-5}
 & Pr-PQ a  & 60/128-128-128-32   & 0.1769     &    0.207    \\
\cline{2-5}
 & Pr-PQ b &    50/32-128-128-32  & 0.1702 & 0.243        \\
\hline
\multirow{7}{*}{DeepDTA-DAVIS} & Pr    & 90            & 0.2425  & 0.2     \\
\cline{2-5}
 & WS   & 2-2-2-2   & 0.2840 & 0.063       \\
\cline{2-5}
 & PQ   &   32-32-2-32    &  0.2567      & 0.237    \\
\cline{2-5}
 & Pr-WS a  &   80/32-2-2-32  &   \textbf{0.2367}     &   0.079     \\
\cline{2-5}
& Pr-WS b &    60/128-2-128-2   &    0.2906      &  0.148      \\
\cline{2-5}
& Pr-PQ a  & 90/128-32-32-32    & 0.2671  & \textbf{0.060}*       \\
\cline{2-5}
& Pr-PQ b &   80/32-2-2-32    &   0.2943       &  0.077       \\
\bottomrule
\end{tabular}
\end{table}

\begin{figure*}[t]
\centering
\scriptsize
\vspace{-0.5cm}
\begin{tabular}{ccc}
\vspace{-0.7cm}
\hspace{-0.5cm}\includegraphics[height=8.05cm,width=0.35\textwidth]{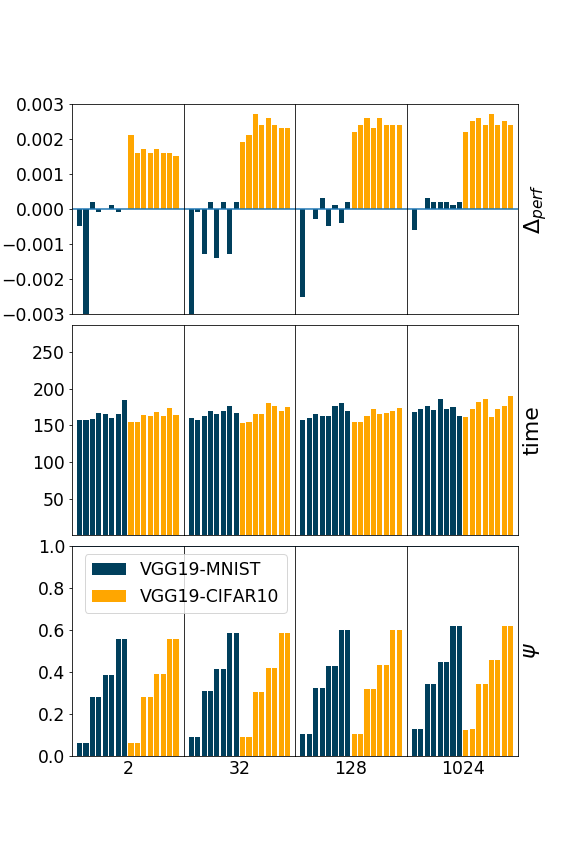}	&
\hspace{-0.5cm}\includegraphics[height=8.05cm,width=0.35\textwidth]{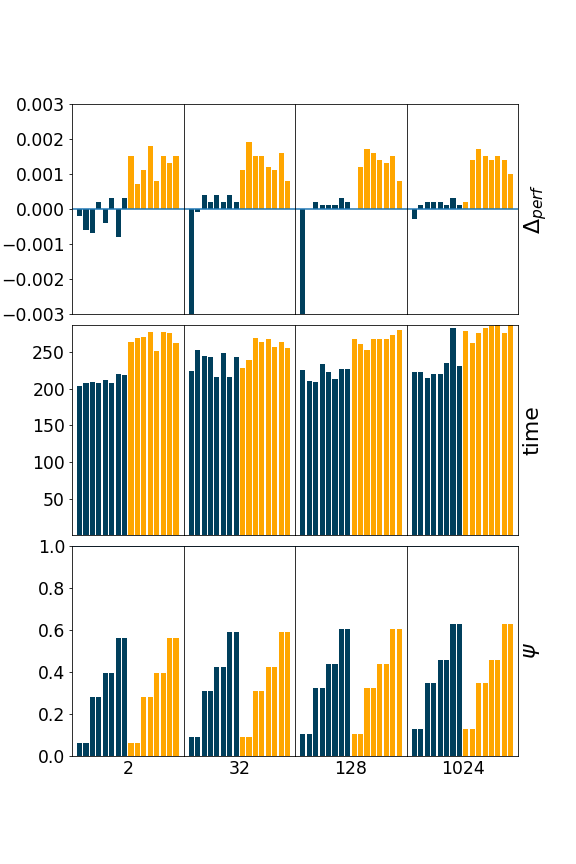} &
\hspace{-0.5cm}\includegraphics[height=8.05cm,width=0.35\textwidth]{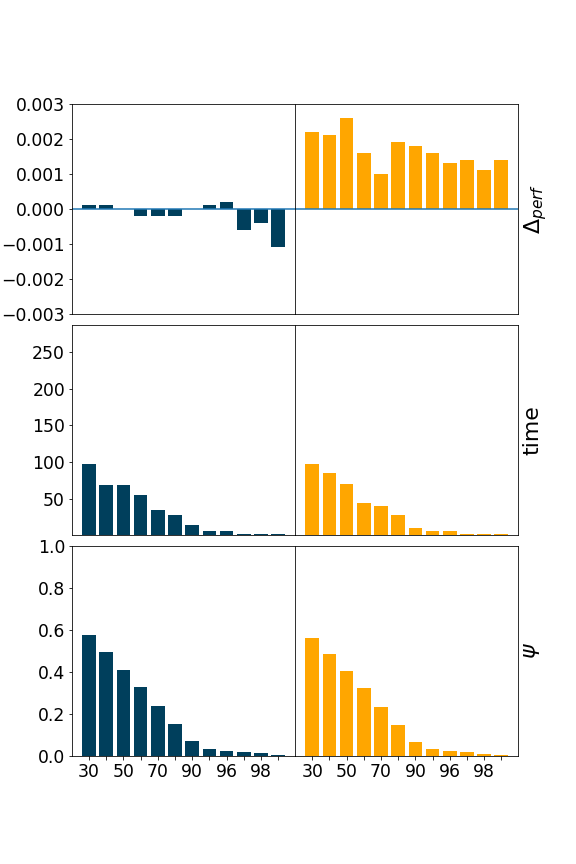}\\
\hspace{-0.5cm}(a) & \hspace{-0.5cm} (b) & \hspace{-0.5cm}(c)\\\vspace{-0.7cm}
\hspace{-0.5cm}\includegraphics[height=8.05cm,width=0.35\textwidth]{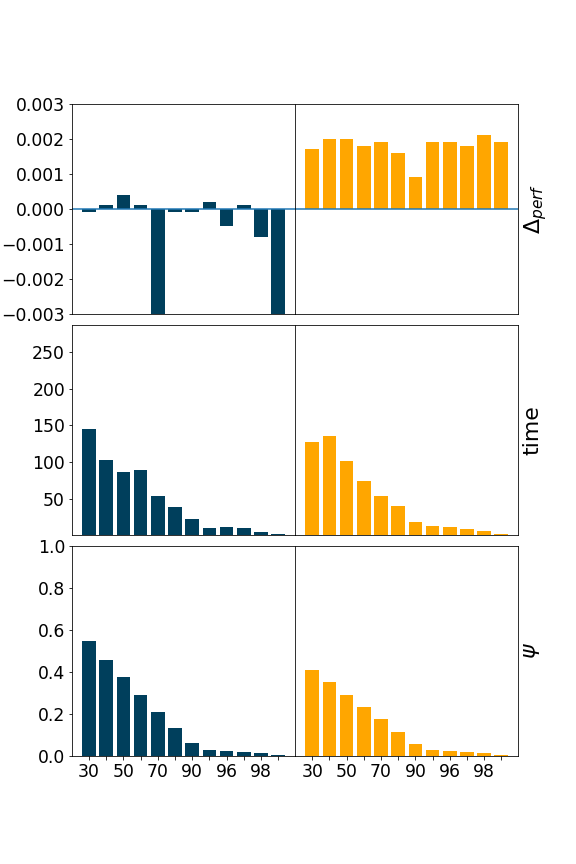}	&
\hspace{-0.5cm}\includegraphics[height=8.05cm,width=0.35\textwidth]{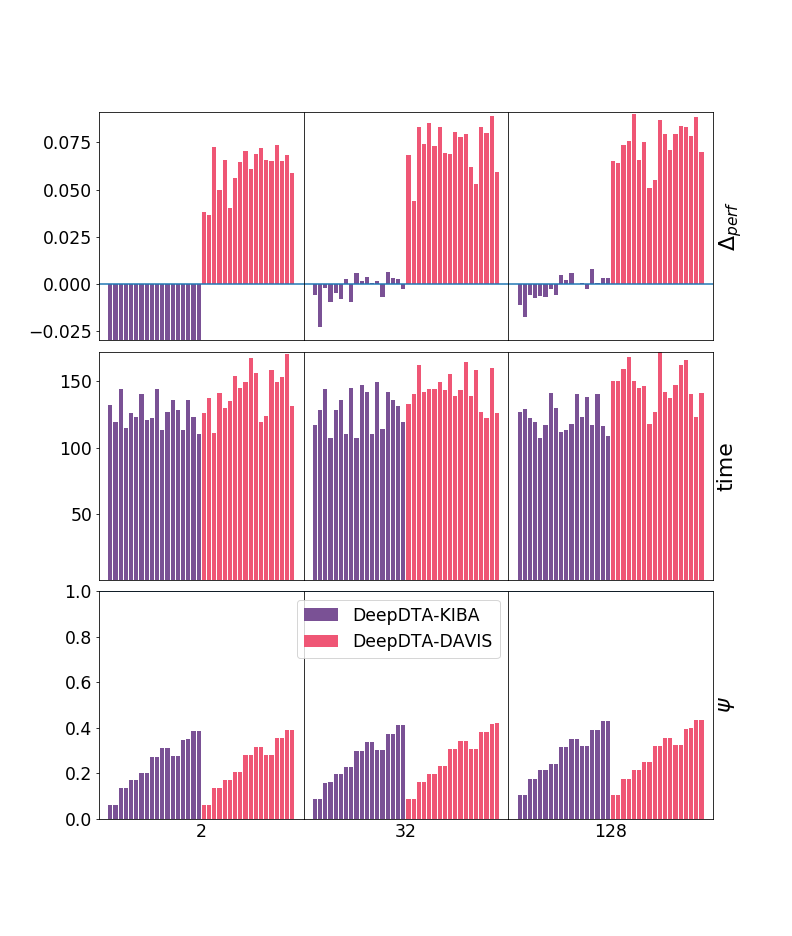} &
\hspace{-0.5cm}\includegraphics[height=8.05cm,width=0.35\textwidth]{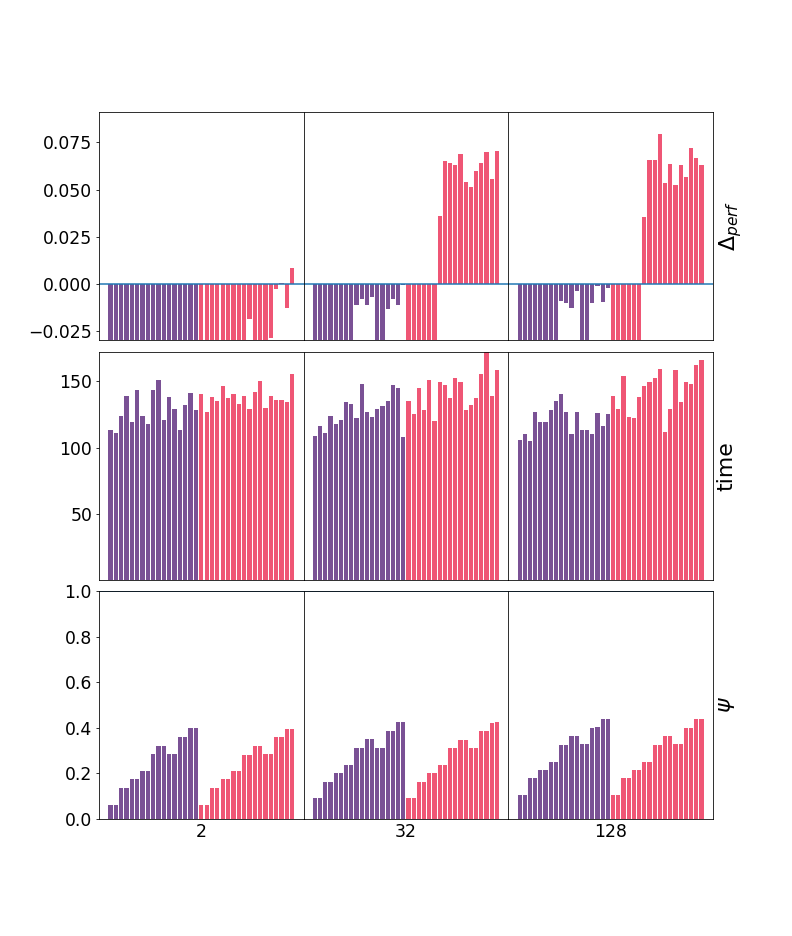}\\
\hspace{-0.5cm}(d) & \hspace{-0.5cm} (e) & \hspace{-0.5cm}(f) \\\vspace{-0.7cm}
\hspace{-0.5cm}\includegraphics[height=8.05cm,width=0.35\textwidth]{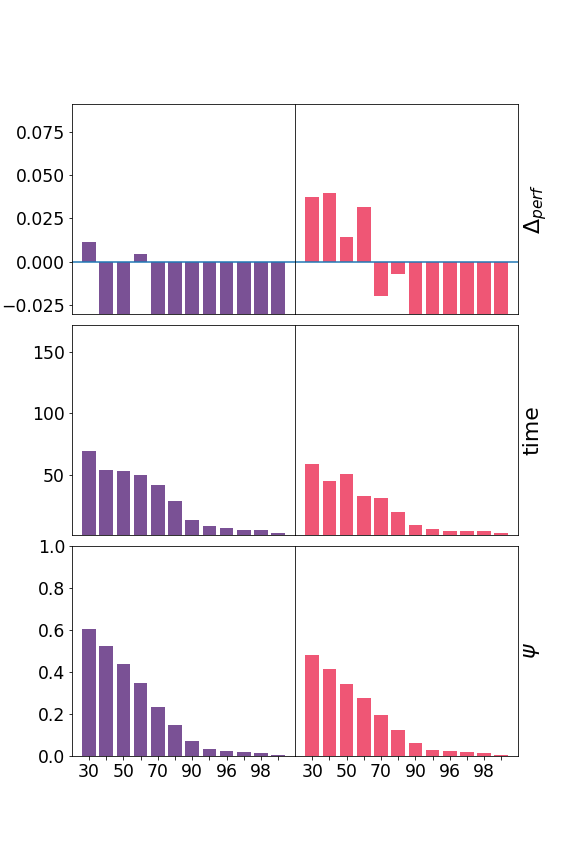}	&
\hspace{-0.5cm}\includegraphics[height=8.05cm,width=0.35\textwidth]{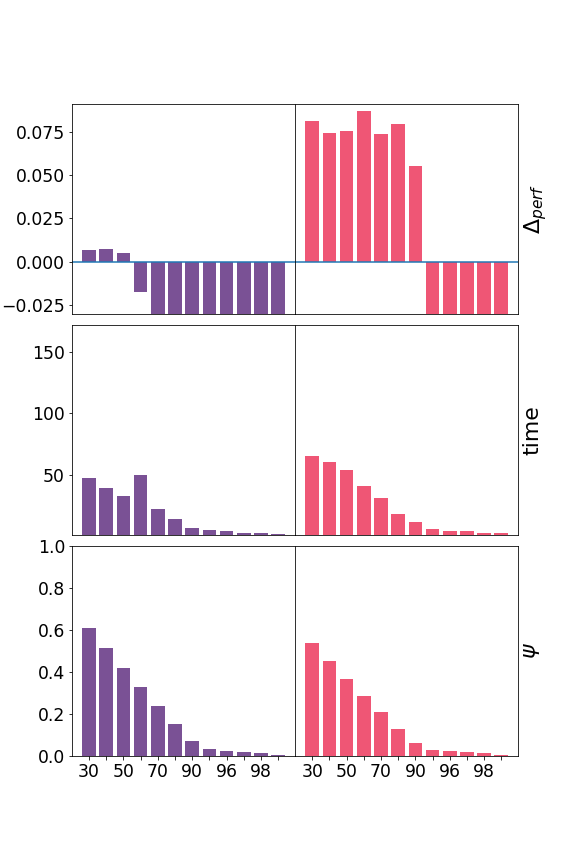} & \\
\hspace{-0.5cm}(g) & \hspace{-0.5cm} (h) &
\end{tabular}
\caption{Overall testing performance for compression methods: a) WS (\HAC\ format), b) PQ (\HAC), c) Pr-WS a (\sHAC), d) Pr-PQ a (\sHAC), e), f), g) and h) the same a), b), c) and  d) methods, but on regression datasets.}\label{fig:all_etrics}
\normalsize
\end{figure*}

To better unveil the behavior of the proposed compression methods and storage formats, in Fig.~\ref{fig:all_etrics} we summarize their testing performance, space occupancy and time ratio, for all tested hyper-parameter configurations.
The \sHAC\ storage format is used, except for techniques producing dense matrices, where \HAC\ is more convenient. Reminding that WS and PQ combinations are reported in increasing order (before all combinations with $k,b=2$ in the first layer, denoted by label $2$, then those with $k,b=32$ in the first layer, label $32$, and so on), on CIFAR-10 and DAVIS most compression techniques outperform the baseline, and this is likely due to overfitting, since on training data they show similar results.
Conversely, on MNIST and KIBA only some compression configurations improve the baseline results, which is however important, since at the same time the compressed model uses much less parameters, confirming results obtained in \cite{Han15}. When using binary quantization ($k,b=2$) clearly we get lower $\psi$ but at the same time worse performance, whereas already with $k,b=32$ the baseline performance is improved on almost all datasets. \sHAC\ occupancy, as expected, gets lower when $p$ increases (and consequently $s$ decreases), along with the time ratio, approaching in turn to $1$ (same testing time). The high time ratios, for some configurations, reflect the fact that the compress {\tt dot} procedure is slower than the {\tt numpy.dot} used by baseline and  leveraging parallel computation. As mentioned in Sect.~\ref{sub:metrics}, the former is still sequential, and we plan to produce a parallel version (cfr.\ Sect.~\ref{sec:sham}).

Comprehensively, a compression technique better that the other ones is not emerging from these results. Nevertheless, a sound result  is that methods providing the lowest occupancy, i.e., those combining weight pruning  and quantization, still achieve high performance; secondly, it seems that the pruning technique is preferable for regression problems, whereas quantization performs better in the setting of classification. However, we believe further studies are necessary to assume this trend as consolidated. Finally, our proposed storage representations, \HAC\ and \sHAC, produce the expected behaviors, being suitable for both dense (\HAC) and sparse (\sHAC) compressed matrices, and remarkably improving the CSC format on sparse matrices.

\section{Conclusions}
This work investigated both classical CNN compression
techniques (like weight pruning and quantization) and a novel probabilistic compression algorithm, combined with a new lossless entropy coding storage of the network. Our results confirmed that model simplification can improve the overall generalization abilities, e.g., due to limited overfitting, and showed that our compressed representation can reduce the space occupancy of the input network, when suitably preceded by pruning and quantization, more than $150$ times.
As meaningful extension of this work, it would be worthy to operate the quantization so as to minimize the entropy of the quantized weights (known as entropy coded scalar/vector quantization), which in turn would lead to shorter entropy coding \cite{Choi20}. In this study indeed we considered them separately,  since the aim was to compare the effectiveness in terms of prediction accuracy of different compression techniques, to detect possible performance trends related to the type of problem. Moreover, other source coding methodologies (known as universal lossless source coding, e.g., the Lempel–Ziv source coding), less sensitive to source statistics, could be applied rather than Huffman coding, being more convenient in practice than the latter, since they do not require the knowledge of source statistics, and having smaller overhead, since the codebook (i.e., dictionary) is built from source symbols while encoding and decoding.




\bibliographystyle{IEEEtran}
\bibliography{references.bib}

\end{document}